\pdfoutput=1

\documentclass{article}

\newcommand{\neuripsoptions}{final}
\usepackage[\neuripsoptions,nonatbib]{neurips_2021}

\usepackage[utf8]{inputenc} %
\usepackage[T1]{fontenc}    %
\usepackage{hyperref}       %
\usepackage{url}            %
\usepackage{booktabs}       %
\usepackage{amsfonts}       %
\usepackage{nicefrac}       %
\usepackage{microtype}      %
\usepackage{xcolor}         %
\usepackage{multirow}
\usepackage{tablefootnote}

\usepackage[sort,numbers]{natbib}
\renewcommand\cite{\citep}	%
\usepackage{multirow}
\usepackage{xspace}
\usepackage[normalem]{ulem}
\usepackage{enumitem}
\usepackage{ifthen}
\usepackage{subcaption}
\usepackage{float}
\usepackage{amssymb}
\usepackage{amsmath}
\usepackage{soul}
\newcommand{\benchmark}{FLEX\xspace}
\newcommand{\fewshot}{few-shot\xspace}
\newcommand{\Fewshot}{Few-shot\xspace}
\newcommand{\FewShot}{Few-Shot\xspace}

\definecolor{brightmaroon}{rgb}{0.76, 0.13, 0.28}
\definecolor{darkgreen}{rgb}{0.0, 0.4, 0.13}

\newcommand\Tstrut{\rule{0pt}{2.5ex}}         %

\newcommand{\uniqa}{UniFew\xspace}
\newcommand{\uniqameta}{UniFew\textsubscript{meta}\xspace}
\newcommand{\unifew}{\uniqa}

\newcommand{\res}[2]{\begin{tabular}[t]{@{}c@{}}${#1}$\\ $^{\pm{#2}}$\end{tabular}}

\newcommand{\D}{\ensuremath{\mathcal{D}}\xspace}
\newcommand{\Y}{\ensuremath{\mathcal{Y}}\xspace}
\newcommand{\E}{\ensuremath{\mathcal{E}}\xspace}
\newcommand{\train}{\textrm{train}}
\newcommand{\test}{\textrm{test}}
\newcommand{\val}{\textrm{val}}

\newcommand{\Dtrain}{\ensuremath{\D_\train}\xspace}
\newcommand{\Dval}{\ensuremath{\D_\val}\xspace}
\newcommand{\Dtest}{\ensuremath{\D_\test}\xspace}
\newcommand{\Etest}{\ensuremath{\E_\test}\xspace}
\newcommand{\Etrain}{\ensuremath{\E_\train}\xspace}
\newcommand{\Eval}{\ensuremath{\E_\val}\xspace}
\newcommand{\Ytest}{\ensuremath{\Y_\test}\xspace}

\usepackage{graphicx}
\usepackage{pdfrender}

\usepackage{colortbl}    %
\usepackage{arydshln}

\newcommand{\ourtitle}{\benchmark: Unifying Evaluation for \FewShot NLP}
\title{\ourtitle}

\newcommand{\ourkeywords}{benchmarks, evaluation, few-shot, zero-shot, nlp, prompt-based models, pretrained language models, statistical analysis}

\newcommand{\ourauthors}{Jonathan Bragg, Arman Cohan, Kyle Lo, Iz Beltagy}
\author{%
    Jonathan Bragg\thanks{Equal contribution}
    \qquad
    Arman Cohan\footnotemark[1]
    \qquad
    Kyle Lo
    \qquad
    Iz Beltagy \vspace{3pt}
    \\
    Allen Institute for AI, Seattle, WA\vspace{2pt}\\
    \small{\texttt{\{jbragg,armanc,kylel,beltagy\}@allenai.org}}
}

\definecolor{darkblue}{rgb}{0, 0, 0.5}
\ifthenelse{\equal{\neuripsoptions}{\empty}}{
    \hypersetup{
        colorlinks=true,
        citecolor=darkblue,
        linkcolor=darkblue,
        urlcolor=darkblue,
        pdfauthor={Anonymized for review},
        pdfkeywords={\ourkeywords},
        pdftitle={\ourtitle}
    }
}{
    \hypersetup{
        colorlinks=true,
        citecolor=darkblue,
        linkcolor=darkblue,
        urlcolor=darkblue,
        pdfauthor={\ourauthors},
        pdfkeywords={\ourkeywords},
        pdftitle={\ourtitle}
    }
}

\begin{document}

\maketitle
\begin{abstract}

\Fewshot NLP research is highly active, yet conducted in disjoint research threads with evaluation suites that 
lack challenging-yet-realistic testing setups and fail to employ careful experimental design.
Consequently, the community does not know which techniques perform best or even if they outperform simple baselines.
In response, we formulate the \benchmark Principles, a set of requirements and best practices for unified, rigorous, valid, and cost-sensitive \fewshot NLP evaluation.
These principles include Sample Size Design, a novel approach to benchmark design that optimizes statistical accuracy and precision while keeping evaluation costs manageable.
Following the principles, we release the \benchmark benchmark,\footnote{Benchmark, leaderboard, and benchmark creation toolkit: \url{https://github.com/allenai/flex}. Apache License 2.0} which includes 
four few-shot transfer settings, zero-shot evaluation, and a public leaderboard that covers  diverse NLP tasks.
In addition, we present \uniqa,\footnote{\Fewshot model: \url{https://github.com/allenai/unifew}. Apache License 2.0
}
a 
prompt-based model for \fewshot learning that unifies pretraining and finetuning prompt formats, eschewing complex machinery of recent prompt-based approaches in adapting downstream task formats to language model pretraining objectives. We demonstrate that despite simplicity, \uniqa achieves results competitive with both popular meta-learning and prompt-based approaches.

\end{abstract}

\section{Introduction}

\Fewshot learning, the
challenge of learning from a small number of examples, is critical for developing efficient, robust 
NLP techniques~\cite{wang2020,yin2020}. 
In recent years, separate threads
of \fewshot NLP research %
have
pursued goals like generalization to new classes~\cite[e.g.,][]{bao2020,gao2019a}, adaptation to new domains %
and tasks%
~\cite[e.g.,][]{bansal-coling20,bansal-emnlp20-self-supervised,dou2019}, and direct application of pretrained language models (LMs)~\cite[e.g.,][]{brown2020,Schick2020ExploitingCQ,Schick2020ItsNJ,gao2020}.
Unfortunately, despite the shared goal of advancing \fewshot NLP techniques, the community does not know which techniques work best or even if they perform better than simple baselines.
Evaluation suites across these research threads are disjoint, lack challenging-yet-realistic testing setups (e.g., class imbalance, variable training set sizes, etc.), and do not employ careful experimental design to ensure accurate and precise evaluation estimates and minimal computational burden.
Prior work in \fewshot learning outside of NLP serves
as a stark warning of the consequences of improper measurement: \citet{dhillon2020} showed that techniques from several years of prior work did not make clear progress due to large overlapping accuracy distributions and, moreover, do not outperform a simple, carefully-tuned baseline.

\textbf{Need for systematic benchmark design} As such, a high-quality benchmark is urgently needed to enable rigorous comparison of techniques across disjoint, highly-active threads of \fewshot NLP research.  But what should such an evaluation suite look like?
Some best practices for evaluation of \fewshot methods have been introduced in the computer vision (CV) literature~\cite{triantafillou2020,dhillon2020} and should be applied to NLP.
However, unifying \fewshot NLP work introduces new challenges.
For example, the benchmark needs to test all types of transfer studied in separate research threads 
to measure progress on new techniques that make gains in each of these important generalization settings (\S\ref{sec:transfer}).
Also, given the importance of zero-shot learning and learning from task descriptions \cite{weller2020,Hase2021WhenCM}, the benchmark needs to include zero-shot episodes and textual labels to enable measuring progress for models that do not use conventional supervised training, including methods that leverage the latent knowledge in pretrained LMs \cite{brown2020,gao2020,Zhao2021CalibrateBU}.
Further, the benchmark must accommodate new, computationally-expensive approaches, without overly reducing the number of evaluation episodes at the expense of statistical accuracy~\cite{gao2020,ye2021,bansal-coling20}.

\textbf{Need for a robust \fewshot model} Recent prompt-based models \cite{brown2020} have shown strong results in \fewshot learning. These models leverage the power of (often large) pretrained language models and adapt the format of downstream tasks to the underlying pretraining objective (e.g., Masked Language Modeling). This way, given the right natural language prompt (and sometimes verbalizers \cite{Schick2020ExploitingCQ} and additional demonstrative examples), the model can quickly fine-tune on the downstream task \cite{Schick2020ExploitingCQ,Schick2020ItsNJ,gao2020,liu2019,lu2021fantastically}. However, adapting task formats to the  underlying (masked) language modeling objectives is not straightforward; such models have been shown to be sensitive to varying choices of the prompt/demonstrations, training settings, hyperparameters, and learning algorithms~\cite{Zhao2021CalibrateBU,perez2021,logan-cutting-down-2021}, often requiring large held out sets and/or complex methods to overcomes such challenges. 
Can models eschew complex prompt engineering by unifying pretraining and downstream task formats?

\begin{table}[t]
    \centering
    \footnotesize
    
\caption{
    Comparison of the \benchmark\ benchmark 
    with closest prior work.
    Our benchmark consists of episodes with variable number of shots in the range [1-5] and with class imbalance. %
    ``No extra test data'' refers to excluding validation data
    from testing tasks, to avoid 
    unfairly advantaging
    models that 
    use
    such data~\cite{perez2021}.
    Our benchmark's number of test episodes is selected to balance statistical accuracy and precision, which suffers in few-episode setups, and compute requirements, which is too costly in many-episode setups (\S\ref{sec:stats}).  %
    } 
    \label{tab:comparison}        
    
    \setlength{\tabcolsep}{3.1pt}
    \renewcommand{\arraystretch}{0.8}
\begin{tabular}{@{}lcccccc>{\columncolor[RGB]{220, 255, 255}}c@{}} 
\toprule
\multicolumn{2}{r}{CrossFit\cite{ye2021}} & LM-BFF\cite{gao2020}  & GPT-3\cite{brown2020}& DS\cite{bao2020}   & SMLMT\cite{bansal-emnlp20-self-supervised} & FewGlue\cite{Schick2020ItsNJ} & \textbf{\benchmark (ours)}  \\ 
\midrule
Class Transfer     & -         & -         & -    & \checkmark  & -  & - & \checkmark    \\
Domain Transfer    & -         & -         & -      & -       & \checkmark    & -    & \checkmark   \\
Task Transfer      & \checkmark & -         & -         & -          & \checkmark    & -    & \checkmark   \\
Pretraining Transfer  & -         & \checkmark & \checkmark & -          & -            & \checkmark   & \checkmark  \\ 
Shots per class    & \{16, 32\} & 16        & variable  & \{1,5\}     & \{4,8,16,32\} & \{total 32\}\tablefootnote{The total number of training shots in each episode, not number of shots per class per episode.}& {{[}1–5]}    \\
Variable shots     & -         & -         & \checkmark & -          & -            & -     & \checkmark \\
Unbalanced         & -         & -         & -         & -          & -            & -     & \checkmark   \\
Textual labels     & \checkmark         & \checkmark         & \checkmark         & - & - & \checkmark   & \checkmark  \\
Zero-shot          & -         & \checkmark & \checkmark & -      & -     & -    & \checkmark     \\ 
No extra test data & -         & -         & -         & \checkmark  & \checkmark    & mixed\tablefootnote{Most users use unlabeled examples, though recently, \citet{tam2021} do not.}         & \checkmark     \\
\# test episodes   & 5         & 5         & 1         & 1000       & 10  & 1    & {90}   \\
Reporting          & avg       & avg, SD   & avg       & avg, SD    & avg, SD      & avg, SD    & {all}\tablefootnote{Average (avg), confidence interval (CI), standard deviation (SD), individual episode metrics}  \\  
\# datasets        & 160       & 16        & 37        & 7          & 18           & 8    & {20}     \\                                                                        
\bottomrule 
\end{tabular}
\end{table}

In this paper, we tackle these key issues by introducing \benchmark---\textbf{F}ew-shot \textbf{L}anguage \textbf{E}valuation across (\textbf{X}) many transfer types---and contributing the following:
\begin{itemize}
    
    \item \benchmark\ Principles (\S\ref{sec:design}), 
    a set of requirements and best practices for \fewshot NLP evaluation
    that enables unified, rigorous, valid, and cost-sensitive measurements.
    \begin{itemize}
        \item 
            Sample Size Design: 
            In support of valid, cost-sensitive measurement, we introduce a novel approach to
        \fewshot sample size design (\S\ref{sec:stats}) that optimizes for a benchmark's statistical accuracy and precision while keeping computational costs accessible to a broad range of researchers.

    \end{itemize}
    
    \item \benchmark\ benchmark (\S\ref{sec:benchmark}), 
    an implementation of the \benchmark Principles. It
    tests across \emph{four} \fewshot transfer settings,\footnote{Prior work evaluated at most two settings.} and includes a public leaderboard for \fewshot NLP that covers 20 datasets across diverse NLP tasks (e.g., NLI, relation classification, entity typing). Table~\ref{tab:comparison} summarizes key differences between \benchmark and other \fewshot NLP evaluation suites. %

    \item \uniqa (\S\ref{sec:model}), a prompt-based model for \fewshot learning in NLP.
    While most existing methods leverage pre-trained LMs for few-shot learning, LM pre-training tasks
    do not closely match 
    natural downstream task formats, requiring complex methods (e.g., extensive prompt-engineering, use of verbalizers, episodic hyperparameter tuning, custom learning algorithms) to make these models work in few-shot setting. Instead, the key idea of our model, \uniqa, is to close the gap between pre-training and fine-tuning formats by posing tasks as multiple-choice QA and using an underlying model that is pre-trained on a similar natural QA task format. This eliminates the need for complexities of adapting downstream tasks to the LM objectives, while resulting in competitive performance with both recent \fewshot and meta-learning methods.

\end{itemize}

To aid similar efforts, our release of \benchmark includes a toolkit for benchmark creation and \fewshot NLP model development, which we used to create the \benchmark\ benchmark and train \unifew.

\section{Background and Related Work}
\label{sec:background}
We first provide background and notation for \fewshot learning and evaluation, then discuss related work in NLP and outside NLP that motivated us to create the \benchmark Principles and benchmark.

\paragraph{\Fewshot background and notation}

Broadly, modern approaches to \fewshot learning are evaluated in a three-phase procedure~\cite{Vinyals2016MatchingNF}.
In the first phase, a general-purpose pretrained model is obtained.
In the subsequent ``meta-training'' phase,%
\footnote{Meta-training may include a ``meta-validation" component, for validating generalization.}
techniques aim to adapt the model to be well-suited for \fewshot generalization.
Finally, a ``meta-testing'' phase evaluates the adapted model in new \fewshot prediction settings.

Let $\D$ be a dataset of $(x,y)$ examples with full label set $\Y_\D$.
From it, we construct three {\em sets} of episodes, corresponding to meta-training, meta-validation, and meta-testing and denoted by \Etrain, \Eval, and \Etest, respectively. Each episode in each of these sets is a \fewshot problem with its own test set and other attributes. 
Formally, each episode $E$ is a tuple $(\Dtrain^E,\Dval^E,\Dtest^E,\Y^E_\D)$, where $\Y^E_\D$ is a sampled subset of labels in $\Y_\D$ and $\D^E_{\train|\val|\test}$ are disjoint sets of examples from $\D$ with labels in $\Y^E_\D$.\footnote{In the \fewshot literature, $\Dtrain^E$ and $\Dtest^E$ are also called the {\em support} and {\em query} sets, and $|\Y^E_\D|$ the {\em way}.} For each episode, the model's objective is to correctly predict labels for examples $\Dtest^E$.  To accomplish this, models make use of labeled examples in $\Dtrain^E$, which is typically configured such that each label $i$ in $\Y^E_\D$ has $K_i^E$ provided examples;
$K_i^E$
is known as the \emph{shot}, and the setting when a class has no examples in $\Dtrain^E$ (i.e., $K_i^E = 0$) is called \emph{zero-shot}.

\paragraph{\Fewshot evaluation in NLP}
\label{sec:transfer}
Research in \fewshot NLP has proceeded in several parallel threads, each focused on a different type of transfer ability~\cite{yin2020}.
Each thread has separate evaluation practices, and the vast majority of \fewshot NLP research has limited evaluation to a single transfer type (see Table \ref{tab:comparison}). Here, we describe these types of transfer and their evaluation practices. 

Following the CV literature~\cite{Vinyals2016MatchingNF,triantafillou2020}, one thread of \fewshot NLP focuses on \textbf{class transfer}, the problem of generalizing from a supervised set of classes at meta-train time to a different set of classes from the same dataset at meta-test time.  Evaluation typically involves splitting classes $\Y_\D$ into $\Y^\D_\train$, $\Y^\D_\val$ and $\Y^\D_\test$ disjoint subsets.  Class transfer has been studied on many text classification tasks~\cite{bao2020}, including  relation classification~\cite{han2018,gao2019a,sun2019}, intent classification~\cite{sun2019,krone2020}, inter alia. In contrast, \textbf{domain transfer} keeps the same classes between meta-training and meta-testing but changes the textual domain (e.g., generalizing from MNLI to science-focused SciTail~\cite{dou2019,bansal-emnlp20-self-supervised}). Evaluation then requires identifying pairs of datasets with the same classes $\Y_\D$, where one dataset's episodes are assigned to \Etrain and the other's to \Etest.  Domain transfer has also been studied on many tasks~\cite{bansal-coling20,bansal-emnlp20-self-supervised}, including dialogue intent detection \& slot tagging~\cite{hou2020}, sentiment classification~\cite{yu2018}, NLI~\cite{dou2019}, and machine translation~\cite{gu2018,sharaf2020}.

Researchers have also begun to study \textbf{task transfer}, the problem of generalizing from a set of tasks at meta-train time to unseen tasks at meta-test time.  Evaluation requires tasks (e.g., NLI) appearing in \Etest\ {\em not} to appear in \Etrain or \Eval. Prior work has used GLUE tasks~\cite{wang2018} for  meta-training before meta-testing on tasks such as entity typing~\cite{bansal-emnlp20-self-supervised,bansal-coling20}, while other work instead used GLUE for meta-testing~\cite{dou2019}. Very recent work has studied task transfer over a large set of datasets~\cite{ye2021,zhong2021adapting}.
A limited amount of work evaluates both domain and task transfer~\citep{dou2019,bansal-coling20,bansal-emnlp20-self-supervised}. An important emerging line of work (not noted by \citet{yin2020}) is \textbf{pretraining transfer}, the problem of whether pretrained language models can perform well at meta-test time without any meta-training.  Evaluation in this setting requires $\Etrain,\Eval=\emptyset$. Prior work has shown that pretrained language models are capable of surprising performance on many \fewshot tasks, even without fine-tuning \cite{brown2020}. More recent work, mainly focusing on text classification, has reported further gains with cloze-style formats~\cite{tam2021,Schick2020ExploitingCQ,Schick2020ItsNJ}, prompt engineering~\cite{gao2020}, or calibration~\cite{Zhao2021CalibrateBU}. \benchmark is designed to exercise all four of these transfer types from previous work.

\paragraph{\Fewshot evaluation outside NLP}
\label{sec:relate-outside}
The \fewshot\ learning literature has largely focused on image classification, with the introduction of increasingly complex meta-learning algorithms
~\cite[e.g.,][]{Vinyals2016MatchingNF,snell2017,rusu2019,lee2019,finn2017}.
However, more recent work has shown that simple fine-tuning baselines are in fact competitive, and attribute this delayed discovery to problematic evaluation methodology~\cite{chen-iclr2019,dhillon2020}. %
\benchmark adopts recommended methodology~\cite{triantafillou2020,dhillon2020}, and we introduce an analogous
baseline (\uniqa) to 
provide a strong measurement foundation for \fewshot NLP.

\section{\benchmark Principles for \FewShot NLP Evaluation}
\label{sec:design}

We now enumerate key desiderata for a \fewshot NLP benchmark capable of solving the urgent problems with \fewshot NLP evaluation, %
including separate evaluations for each transfer type %
and failure to incorporate best measurement practices from other domains
(\S\ref{sec:relate-outside}).

\textbf{Diversity of transfer types}
To make NLP models broadly useful, \fewshot NLP techniques must be capable of class, domain, and task transfer.
Moreover, techniques should make use of the relevant supervision provided during meta-training to increase performance compared to the pretraining transfer setting.
The benchmark should measure all four transfer settings to ensure that the community develops techniques that improve on strong pretraining transfer baselines, and enable comparison across these currently separate threads of research.

\textbf{Variable number of shots and classes}
To better simulate a variety of real-world scenarios, the benchmark should include a variety of training set sizes and numbers of classes~\cite{triantafillou2020}.
Testing robustness to these factors is crucial;
\fewshot techniques are often sensitive to changes in these factors~\cite{cao2020}, yet all prior \fewshot NLP evaluations we are aware of used a fixed number of training shots and classes, known in advance during meta-training.

\textbf{Unbalanced training sets} 
The benchmark should also include unbalanced training sets with different training shots per class, another realistic setting adopted by CV benchmarks~\cite{triantafillou2020}.
Class imbalance has also been observed to degrade performance~\cite{Buda2018ASS,ochal2021}, yet prior \fewshot NLP evaluations do not include this setting either.

\textbf{Textual labels} 
While numerical label values are often used in classification tasks, descriptive textual labels are also present for many tasks.
Making these textual labels available for use by \fewshot techniques enables the development of techniques that can leverage the class name, like in-context learning~\cite{brown2020}, template generation~\cite{gao2020}, and meta-learning~\cite{luo-etal-2021-dont}.
Textual labels are crucial in particular for zero-shot evaluation.%

\textbf{Zero-shot evaluation} 
We believe zero-shot evaluation is integral to the goals of \fewshot evaluation. 
Similar to the motivation for measuring pretraining transfer, zero-shot evaluation is an important use case and also provides a strong baseline for some tasks.
In the absence of training examples, textual class labels or richer task descriptions~\cite{weller2020} must be provided.
Some recent \fewshot NLP work~\citep[e.g.,][]{brown2020,gao2020} 
evaluated with zero training shots, but most~\cite[e.g.,][]{bao2020,bansal-coling20,ye2021} did not.

\textbf{No extra meta-testing data}
We believe the benchmark should {\em not} provide validation data ($\Dval^E=\emptyset, \forall E\in\Etest$) or unlabeled data for meta-testing tasks,
since \fewshot learning seeks to enable high performance in environments where collecting additional data is costly.%
\footnote{Unlabeled data collection can be costly too, e.g. due to manual filtering~\cite{clark2019}.}
Variation in these dimensions in prior NLP work makes comparison of results extremely difficult
because it is often under-reported and gives unfair advantage to approaches that leverage such data~\cite{perez2021}. For example, per-episode hyperparameter tuning on extra data has been shown to greatly inflate evaluation scores~\cite{gao2020}. A few researchers~\cite{tam2021,bao2020} follow our suggested approach, but others have used many different settings,
from validation sets of various sizes~\cite{brown2020,gao2020,zheng2021fewnlu} to no validation set but a large set of unlabeled examples~\cite{Schick2020ExploitingCQ,Schick2020ItsNJ}.

\textbf{Principled sample size design}
Promising \fewshot techniques can incur significant computational cost per episode, e.g., due to fine-tuning model parameters~\cite{bansal-emnlp20-self-supervised}, searching for prompts~\cite{gao2020}, inter alia. To alleviate these costs, related works often evaluate with a limited number of episodes, which precludes statistically accurate or precise performance estimates.
We believe the benchmark's test sample size should be optimized to enable proper performance evaluation for such techniques, 
while ensuring the computational burden
is inclusive toward researchers without 
large compute resources.

\textbf{Proper reporting of CIs, SDs, and individual results} 
The benchmark should report confidence intervals (CIs) of performance estimates %
and follow recent guidelines~\cite{dhillon2020} to report standard deviations (SDs) for understanding variability.
Moreover, we newly advocate for controlling for the {\em same}
sampled \fewshot episodes across all methods 
and reporting individual episode results, so that researchers can run higher-powered paired statistical tests when comparing results~\cite{dror2018}, crucial when the benchmark has been optimized for low evaluation budgets.

\section{\benchmark Benchmark}
\label{sec:benchmark}

The \benchmark\ benchmark is a unifying, rigorous evaluation suite for \fewshot learning in NLP, which implements the desiderata outlined in the previous section. 
In this section, we describe detailed design decisions and
our accompanying \fewshot NLP toolkit (\S\ref{sec:framework}), which we are releasing to facilitate easily adding NLP datasets and advanced sampling options to future benchmarks.
We also describe the \benchmark leaderboard (\S\ref{sec:leaderboard}).

\subsection{Task and Dataset Selection}
\label{sec:format}

Following GLUE~\cite{wang2018} and other prior work~\cite{bansal-coling20,bao2020,gao2020,Zhao2021CalibrateBU}, we focus on tasks formatted as classification.
Despite recent advances, NLP state-of-the-art models remain significantly worse than human performance on many text classification tasks, particularly in the \fewshot setting. Automatic scoring of classification tasks is also more reliable than text generation tasks.

We selected datasets across three recent \fewshot NLP evaluation suites, which separately studied class transfer~\citep{bao2020}, domain and task transfer~\cite{bansal-coling20,bansal-emnlp20-self-supervised}, and pretraining transfer~\citep{gao2020}.
Our benchmark includes a broad mix of  tasks (NLI, question classification, entity typing, relation classification, and sentiment analysis) and formats (document, sentence, sentence pair). More complete dataset and license details are available in the following subsection and Appendix~\ref{sec:appendix:dataset}.

\subsection{Meta-Evaluation Protocols}
\label{sec:training-protocols}

As discussed earlier, \benchmark evaluates four different types of transfer: Class, Domain, Task, and Pretraining Transfer.
To support all types, we report results to the \benchmark benchmark both \emph{without} meta-training (pretraining-only) and \emph{with} meta-training. 
This reporting scheme evaluates the performance  
of the basic pretrained model and the benefit (or lack thereof) of meta-training. A similar reporting scheme was proposed by~\citet{triantafillou2020} for CV.

\textbf{Pretraining-Only}
In this setting, the pretrained model is directly meta-tested on our benchmark without any additional training. This is the \emph{Pretraining Transfer} setting, and it is the most difficult, but given the recent success of pretrained models in NLP for \fewshot learning~\cite{brown2020,gao2020}, we believe that comparison to models without any meta-training is important for NLP tasks.

\textbf{Meta-Trained}
In this setting, the model is meta-trained then meta-tested on our benchmark.
We carefully selected and split datasets across meta-train/validation/test  in order to enable testing of Class, Domain, and Task transfer with a single meta-training phase (to reduce computational burden).
Datasets involved in each transfer setting (detailed split information in Table~\ref{tab:dataset} in Appendix~\ref{sec:appendix:dataset}):

\begin{itemize}
    \item \emph{Class Transfer}: FewRel~\cite{han2018}, HuffPost~\cite{misra2018}, Amazon~\cite{he2016}, 20News~\cite{lang1995}, and Reuters~\cite{lewis1997} take part in meta-training and
    meta-testing but with different classes.
    \item \emph{Domain Transfer}: MR~\cite{pang2005}, CR~\cite{hu2004}, SNLI~\cite{bowman2015}, and SciTail~\cite{khot2018} are only in the meta-testing phase, but the corresponding sentiment and NLI datasets exist in the meta-training phase (MNLI~\cite{williams2018}, QNLI~\cite{rajpurkar2016}, and SST-2~\cite{socher2013}).
    \item \emph{Task Transfer}: Subj~\cite{pang2004}, TREC~\cite{voorhees2000}, and CoNLL~\cite{sang2003} are also for meta-testing only, and they represent tasks that the model does not encounter during meta-training.
\end{itemize}

Instead of per-episode hyperparameter tuning,
we provide meta-validation episodes \Eval for learning (during meta-training) global hyperparameters that work across all episodes.
Specifically, the meta-validation dataset splits (see Table~\ref{tab:dataset}) consist of CoLa~\cite{warstadt2019} for task transfer, WNLI~\cite{levesque2011} for domain transfer, and the validation splits used by \citet{bao2020} for all class transfer datasets.
Following \citep{bansal-coling20}, we also include meta-training datasets MRPC~\cite{dolan2005}, RTE~\citep{dagan2005,bar-haim2006,giampiccolo2007,bentivogli2009}, and QQP~\cite{wang2018}.

\subsection{Episode Sampling}

We describe how our benchmark samples meta-testing episodes \Etest. For meta-training, we allow users to sample from \Etrain, \Eval in any way,
or directly use the underlying dataset splits. %

\textbf{Number of classes}
For Class Transfer datasets, \benchmark evaluates model robustness to variable number of new classes.
When constructing episode $E$ from one of these datasets $\D$, our benchmark samples an episode-specific number of classes from dataset $D$, the sampler picks a random number from the range $\Y_\D^E \sim \mathrm{Unif}(5, \min(|\Y_\D|, 10))$.%
\footnote{We limit to 10 classes to avoid undue burden on in-context approaches that fit examples in memory~\cite{brown2020}, and use a lower bound of 5 classes to match prior work~\cite{bao2020}.}
For Domain and Task Transfer, the number of classes is fixed to the maximum number of classes in each dataset
because Class Transfer is not being evaluated.

\textbf{Number of shots}
Following prior work outside NLP~\cite{triantafillou2020,ochal2021}, our benchmark samples the training shot independently for each episode $E$ and class $i$, as $K^E_i \sim \mathrm{Unif}(K_\mathrm{min},K_\mathrm{max})$, where $K_\mathrm{min}=1$.
Given strong performance of NLP models with few or even zero examples~\cite{brown2020,weller2020} and following prior work~\citep{bao2020}, we set the limit $K_\mathrm{max}=5$.
Separately, we allocate an equal number of episodes as zero-shot, where we instead set $\Dtrain^E=\emptyset$ (equivalently, $K_i^E=0, \forall i$).
In each episode, examples are sampled uniformly at random without replacement (but can be reused across episodes).\footnote{These samples represent an unbiased performance estimate, but do not eliminate underlying dataset biases.} 
Following \citet{triantafillou2020}, we select a testing shot that is balanced across classes and leaves roughly half of examples for sampling the training examples.
The total number of episodes for each reported configuration (pair of dataset and either zero- or few-shot) is set to 90 using Sample Size Design (\S\ref{sec:stats}).

\subsection{Extensible Toolkit for Benchmark Creation and Model Training \& Evaluation}
\label{sec:framework}
Alongside the \benchmark benchmark, we release an extensible, highly-configurable Python toolkit, which we used to generate the benchmark%
, and train and evaluate our models.
Unlike existing meta-learning frameworks (e.g., Torchmeta \cite{deleu2019torchmeta}, learn2learn \cite{Arnold2020-ss}),
our framework makes available a wide range of community-contributed NLP datasets and utilities via 
 HuggingFace Datasets \cite{2020HuggingFace-datasets}.\footnote{Apache License 2.0. Full license details for all software dependencies available in Appendix~\ref{sec:appendix:license}.}
Our code also provides advanced sampling utilities (e.g., for class imbalance), ensures reproducibility by checksumming generated episodes, and reports all recommended statistics.

\subsection{Public Leaderboard}
\label{sec:leaderboard}

We provide public leaderboards for each of the meta-evaluation protocols: Pretraining-Only\footnote{\url{https://leaderboard.allenai.org/flex/}} and Meta-Trained.\footnote{\url{https://leaderboard.allenai.org/flex_meta/}}
Submissions take the form of a text label predictions file, which is produced by our toolkit.
Results are reported with confidence intervals, standard deviations, and individual predictions on request.
See Appendix~\ref{sec:appendix:leaderboard} for a screenshot of the results interface.

\section{Sample Size Design: Balancing Statistical Measurement \& Compute Cost}
\label{sec:stats}

We demonstrate a principled approach to determining the optimal sample size configuration in our \fewshot benchmark.
A proper benchmark should produce performance estimates that are \emph{accurate}, close to the true value, and \emph{precise}, low variance. A large (test) sample size can achieve this, yet must be considered alongside computational cost so that a broad community of researchers with differing amounts of compute resources can participate. This decision is further complicated in the \fewshot setting, where sample size refers to both the number of test episodes $|\Etest|$ and the number of test examples $|\Dtest^E|$ per episode $E \in \Etest$. For practicality, we consider $\overline{|\Dtest|}$, the mean $|\Dtest^E|$ across all episodes, rather than every $|\Dtest^E|$.
It remains unknown how one should best distribute test examples between $|\Etest|$ and $\overline{|\Dtest|}$: More episodes each with fewer examples, or fewer episodes each with many examples?  Prior work has been inconsistent in this regard.
For example, \citet{gao2020} used $|\Etest|=5$ and large $\overline{|\Dtest|}$,  while \citet{bao2020} used $|\Etest|=1000$ and much smaller $\overline{|\Dtest|}$.

\begin{figure}[t]
    \centering
    \begin{subfigure}[t]{0.5\textwidth}
        \centering
        \includegraphics[width=0.75\linewidth]{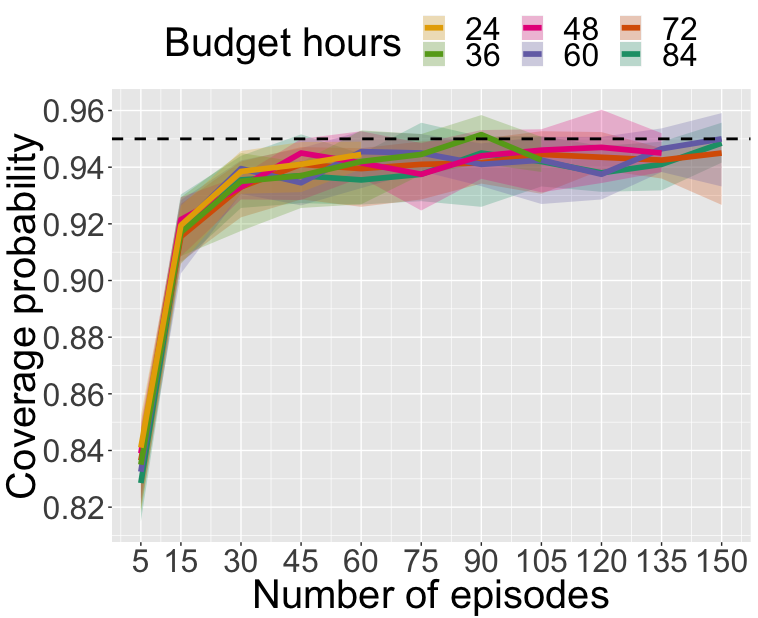}    
        \caption{Coverage probability of 95\% CIs.}
        \label{fig:power-a}
    \end{subfigure}%
    ~ 
    \begin{subfigure}[t]{0.5\textwidth}
        \centering
        \includegraphics[width=0.75\linewidth]{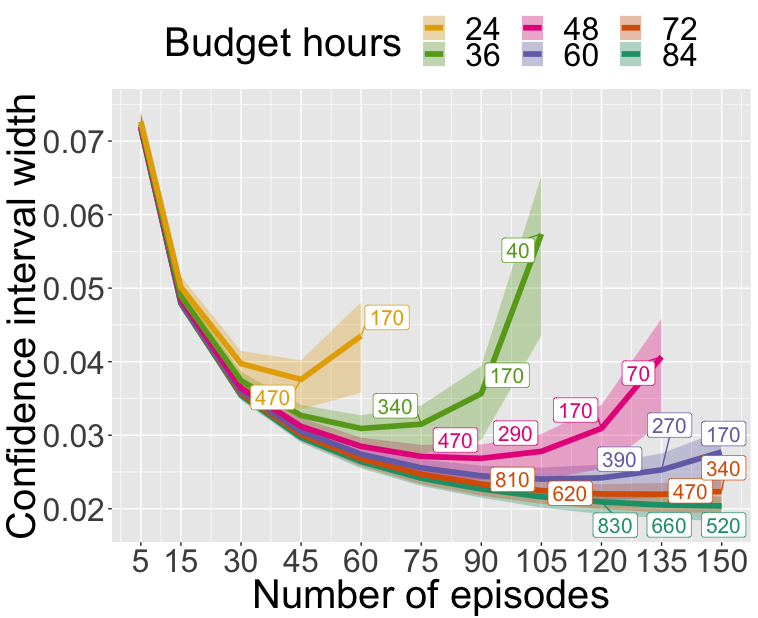}    
        \caption{Mean width of 95\% CIs.}
        \label{fig:power-b}
    \end{subfigure}
    \caption{Results of simulation study described in \S\ref{sec:stats}. Each curve corresponds to a compute budget constraint $C$ (GPU-hours). Each point on a curve is an allocation of test data between the number of test episodes $|\Etest|$ or mean number of examples per episode $\overline{|\Dtest|}$ such that evaluation can be completed within given budget. Per curve, lower values of $|\Etest|$ correspond linearly to larger values of $\overline{|\Dtest|}$, which are shown as numerical text annotations in (b).  Error bars represent the 10$^{th}$ and 90$^{th}$ percentile values from repeated simulations across $\mu_{acc} \in \{0.3, 0.35, \dots, 0.95\}$.
    }
    \label{fig:power}   
\end{figure}

Inspired by simulation techniques for informing statistically-powered experimental design~\citep{Card2020WithLP}, we study how different configurations of $|\Etest|$ and $\overline{|\Dtest|}$ across different compute budgets $C$ impact the accuracy and precision of our estimated CIs, specifically with respect to \emph{coverage probability}~\cite{rubin-schenker-confidence-interval-coverage} and \emph{width}.
First, we estimate per-episode and per-test-example costs of our \fewshot model (\S\ref{sec:model}) to obtain valid $(C, |\Etest|,\overline{|\Dtest|})$ configurations s.t. the full benchmark completes within given $C$ (GPU-hours).\footnote{Costs estimated using a Quadro RTX-8000 GPU with 48Gb memory. For few-shot settings, model was trained with 300 steps. Per-episode and per-test-example costs were approx. 95--98 and 0.7--0.11 GPU-sec, respectively. Using a model with high per-episode cost for this analysis allows us to define a lower-bound sample size requirement; we can always test inexpensive or zero-shot models on more $|\Etest|$ or $\overline{\Dtest}$ within budget.}  Then, for each $(C, |\Etest|,\overline{|\Dtest|})$, we perform 1000 simulation runs, in which each run samples predictions under a true model accuracy $\mu_{acc}$ and computes a single 95\% CI, its width, and whether it correctly covers $\mu_{acc}$.  Averaging over simulation runs gives us estimates for the coverage probability and width of our benchmark's CI for a single $(C, |\Etest|,\overline{|\Dtest|})$.  We repeat this whole procedure for different $\mu_{acc} \in \{0.3, 0.35, \dots, 0.95\}$ to cover a wide range of possible model performances observed across many datasets (see Table~\ref{tab:res}).

Figure~\ref{fig:power} shows CI coverage probability and width for many $(C, |\Etest|,\overline{|\Dtest|})$ configurations.  First, we find in Figure~\ref{fig:power-a} that sufficiently-many test episodes (i.e., $|\Etest| > 60$) is needed to guarantee coverage probability of our CIs is within one percentage point of the target 95\%, a trend that holds regardless of compute budget.
Small $|\Etest|$ also corresponds to large CI widths across all considered budgets in Figure~\ref{fig:power-b}.  This suggests that the choices of $|\Etest|=1,5,10$ in prior work~\cite{gao2020,ye2021,bansal-emnlp20-self-supervised,Schick2020ItsNJ}
can mean inaccurate and wide CIs, while choices of $|\Etest|=1000$~\cite{bao2020} can be prohibitively costly for methods with high %
training cost.

Next, Figure~\ref{fig:power-b} reveals (i) diminishing returns in CI width (decrease in $y$-axis) as compute increases, and (ii) existence of an optimal balance between $|\Etest|$ and $\overline{|\Dtest|}$ for each budget.
Restricting our consideration to budgets with optima satisfying sufficient coverage probability ($|\Etest| > 60$), the minimum viable budget is 36 GPU-hours.  Then, assessing the marginal benefit of each 12 GPU-hour budget increase in terms of marginal reduction in CI width between optima, we arrive at our \benchmark configuration of $|\Etest| = 90$ and $\overline{|\Dtest|} \approx 470$ under a budget of $C=48$ GPU-hours.\footnote{Consider budget increases $36 \to 48$, $48 \to 60$, $60 \to 72$ and $72 \to 80$.  The first reduces CI width by 13\%.  Further increases reduce CI width by an additional 9\%, 7\%, and 5\%, respectively. We choose $C=48$ based on these diminishing returns. %
} Further details are in Appendix~\ref{sec:appendix:simulation}.

\section{\uniqa: A \FewShot Learning Model by Unifying Pre-training and Downstream Task Formats}
\label{sec:model}

Despite their encouraging results, existing works on few-shot learning in NLP are based on either customized and often complex meta-learning algorithms \cite{si2018learning,bansal-coling20,bansal-emnlp20-self-supervised,bao2020}, heavy manual/automated engineering of textual descriptions or prompts \cite{Schick2020ExploitingCQ,Shin2020AutoPromptEK,gao2020,Zhao2021CalibrateBU}, ordering of training examples \cite{Schick2020ItsNJ,lu2021fantastically}, extensive hyperparameter tuning on held-out sets \cite{Schick2020ExploitingCQ,gao2020,lu2021fantastically}, or custom learning algorithms \cite{Schick2020ExploitingCQ,tam2021}.
We present \uniqa, a strong \fewshot learning model across \emph{all} transfer settings and datasets tested, that eschews the need for incorporating the above-mentioned complexities and challenges.

\uniqa is a prompt-based model \cite{Schick2020ItsNJ}, a class of models that tailor the input/output format of their data to match the format used during pretraining.  While this technique allows them to perform a task without the need for additional classification layers, prompt-based models are typically sensitive to the choice of the prompts, which can require extensive search, trial-and-error, and even additional models to get right~\cite{gao2020,Zhao2021CalibrateBU}.
To avoid this issue while still leveraging the strong capabilities of pretrained models, \uniqa (1)
converts examples into multiple-choice question-answer (QA) format, and (2) uses UnifiedQA~\cite{Khashabi2020UnifiedQACF}, a T5~\cite{Raffel2020ExploringTL} model further pretrained on a large collection of QA pairs.\footnote{UnifiedQA and T5 both use Apache License 2.0. We use publicly-released large-size model weights.}\textsuperscript{,}\footnote{None of the supervised datasets 
in the pretraining of UnifiedQA or T5 are in \benchmark.}

Compared to other prompt-based models, \uniqa has two main strengths. First, the prompt design problem is much simpler because UnifiedQA questions had well-defined formats. For example, we only need four general prompt templates which cover
all 20 datasets in the \benchmark benchmark, while prior works have needed specialized prompts for each dataset. Second, UnifiedQA's multiple-choice format ensures the model outputs a valid class label, without the need for learned or manually-defined mappings or verbalizers 
required for other prompt-based methods~\cite{gao2020,Schick2020ExploitingCQ}.\footnote{In rare cases, especially for zero-shot, UnifiedQA may generate an invalid answer (e.g., ``Yes, Yes, No'' instead of ``Yes''). We use simple heuristics to normalize the answer in such cases.} 
In concurrent work, \citet{zhong2021adapting} also show the benefit of performing meta-tuning on a variety of datasets; while their task setup as Q/A is similar to \uniqa, they focus exclusively on binary zero-shot classification tasks and, unlike \uniqa, do not handle multi-class or \fewshot problems.

We experiment with \uniqa both without and with meta-training on the \benchmark benchmark's meta-training data, following the \benchmark protocol (\S\ref{sec:training-protocols}). We call the meta-trained variant \uniqameta. We use simple prompts in the format of question followed by choices followed by the answer (according to the UnifiedQA original format). The exact prompts used are provided in Appendix \ref{sec:prompts}.

\textbf{Training details\hspace{1em}} For meta-training and meta-validation of \uniqa, we sampled \Etrain and \Eval with 5-class, 5-training-shot sampling with the same number of shots per class.\footnote{Users of \benchmark can specify the sampling configuration of \Etrain and \Eval as desired.} %
We trained the model for total number of 30K steps, using a linear learning rate scheduler with peak rate of $3e{-}5$, 200 warmup steps, and batch size of 4; we selected the best checkpoint based on \Eval performance. At meta-test time, for each episode, we trained the model on the episode's training examples (if they exist) and predicted the outputs on test examples.
For training at meta-test time, we used constant learning rate of $3e{-}5$ and batch size of 4, and trained the model for 400 steps.\footnote{For comparison with \cite{gao2020} we trained the model for 600 steps.} We used NVidia RTX8000 GPUs, which take about 7 GPU-hours for meta-training and 48 GPU-hours for meta-testing. For meta-testing we split the episodes among 8 GPUs to speed up evaluations.

\section{Experiments}
\label{sec:compare-results}

\paragraph{Comparing \uniqa with prior work}
To demonstrate the efficacy of \uniqa, we evaluate it against state-of-the-art approaches for few-shot and meta-learning in NLP: LM-BFF \cite{gao2020}, a language model prompt-based fine-tuning method, as well as Distributional Signatures (DS) \cite{bao2020} and H-SMLMT \cite{bansal-emnlp20-self-supervised}, two state-of-the-art meta-learning techniques. Refer to Appendix \ref{sec:appendix:baselines} for details on these methods.

\begin{table}
\renewcommand{\arraystretch}{0.56}
    \centering
    \small
    \caption{Comparing \uniqa with prior methods on their respective test suites, reporting mean accuracy (and standard deviation). 
    For each test suite, for each result set on same number of shots, we indicate with $\rhd$ when results are directly comparable: (i) either both use meta-training (H-SMLMT \& DS with \uniqameta) or neither do (LM-BFF with \uniqa). We \textbf{bold} the better of the two. 
    }
    \label{tab:res-compare}    
    \begin{subtable}[t]{0.36\textwidth}
    \setlength{\tabcolsep}{2.2pt}
        \caption{H-SMLMT~(\citet{bansal-emnlp20-self-supervised}) \hbox{}\hspace{2.5em}\phantom{.}}
    \label{tab:bansal}
    \renewcommand{\arraystretch}{0.76}
    \begin{tabular}[t]{@{}llccr@{}}
        \toprule
        \multicolumn{2}{l}{Model} & Shots &  CNLL & SciT \\
        \midrule
        
        $\rhd$ & H-SMLMT & 4 & \res{57.6}{7.1} & \res{76.8}{3.4} \\
        & \uniqa & 4 & \res{76.6}{2.6} & \res{65.1}{9.9} \\
        $\rhd$ & \uniqameta & 4 & \res{\bf{79.7}}{2.8} &  \res{\bf{85.4}}{2.5} \\ \hdashline[1pt/1pt]
        $\rhd$ & \Tstrut H-SMLMT 	 & 8 & \res{70.2}{3.0} & \res{79.1}{1.1}  \\
        & \uniqa & 8 & \res{80.6}{3.7} & \res{70.9}{5.2} \\
        $\rhd$ & \uniqameta & 8 & \res{\bf{81.2}}{3.8}& \res{\bf{86.8}}{1.4} \\ \hdashline[1pt/1pt]
        $\rhd$ & \Tstrut H-SMLMT & 16 & \res{80.6}{2.8} &  \res{80.4}{1.4}  \\
        & \uniqa & 16 & \res{85.8}{1.9} & \res{76.7}{4.6} \\
        $\rhd$ & \uniqameta & 16 & \res{\bf{87.9}}{1.9} & \res{\bf{85.4}}{2.5} \\
        \bottomrule
    \end{tabular}
    \end{subtable}
    \begin{subtable}[t]{0.63\textwidth}
    \setlength{\tabcolsep}{3.8pt}
    \centering
    \caption{LM-BFF~(\citet{gao2020})}
    \label{tab:gao}    
    \begin{tabular}[t]{@{}llcccccr@{}}
    
    \toprule
        \multicolumn{2}{l}{Model} & Shots &  CR & MR & SNLI & Subj & TREC \\
        \midrule
        $\rhd$ & LM-BFF$_\textrm{man}$\tablefootnote{\citet{gao2020}'s automatic prompt search and in-context learning are not available in the zero-shot setting, so they instead use manually-designed prompts.} & 0\tablefootnote{Zero-shot results from \citet{gao2020} are on the entire test set, so there is no reported standard deviation.} & \textbf{79.5} & \textbf{80.8} &  49.5 & \textbf{51.4} & \textbf{32.0}   \\
        $\rhd$ & \Tstrut\uniqa & 0 & 78.8 & 74.8 &  \textbf{54.4} & 50.3 & 15.0  \\
        & \Tstrut\uniqameta & 0 & 92.1 & 90.5 & 83.8 & 56.8 & 39.1  \\
        \hdashline[1pt/1pt]\Tstrut
        $\rhd$ & LM-BFF & 16/16\tablefootnote{16/16 denotes 16 shots for training plus 16 more for validation which we only use for early stopping while \citet{gao2020} use for grid-search hyperparameter tuning.} & \res{91.0}{0.9} & \res{\bf{87.7}}{1.4} & \res{\bf{77.5}}{3.5} & \res{\bf{91.4}}{1.8} & \res{\bf{89.4}}{1.7}   \\
        $\rhd$ & \uniqa & 16/16 & \res{\bf{92.2}}{0.8} & \res{87.2}{0.1} &  \res{75.6}{1.5} & \res{84.6}{5.4} & \res{86.7}{0.3}  \\ 
        & \uniqameta & 16/16 & \res{92.7}{0.4} & \res{90.2}{0.8} & \res{84.9}{0.5} & \res{87.6}{2.0} & \res{86.1}{0.4}  \\        
        \bottomrule
    \end{tabular}
    \caption{Distributional Signature~(\citet{bao2020})}
    \label{tab:bao}
    \begin{tabular}{@{}llcccccr@{}}
    \toprule
        \multicolumn{2}{l}{Model}  & Shots & Amzn$^\dag$ & FRel$^\dag$ & HuffP$^\dag$ & 20N$^\dag$ & Reut$^\dag$ \\ 
        \midrule

            $\rhd$ & DS & 1 & \res{62.7}{0.7} & \res{67.1}{0.9} & \res{43.1}{0.2} & \res{52.2}{0.7} & \res{81.8}{1.6}  \\
            & \uniqa & 1 & \res{82.1}{8.5} & \res{75.7}{13.2} & \res{65.9}{13.4} & \res{58.4}{11.6} & \res{92.0}{8.3} \\
            $\rhd$ & \uniqameta & 1 & \res{\bf{84.3}}{8.9} & \res{\bf{90.6}}{6.2} & \res{\bf{78.6}}{6.9} & \res{\bf{70.3}}{9.1} & \res{\bf{96.9}}{2.5} \\ 
            \hdashline[1pt/1pt]
            $\rhd$ & \Tstrut DS & 5 & \res{81.2}{0.3} & \res{83.5}{0.3} & \res{63.5}{0.1} & \res{68.3}{0.2} & \res{96.0}{0.3} \\ 
            & \uniqa & 5 & \res{88.5}{7.4} & \res{88.8}{6.5} & \res{77.1}{6.0} & \res{72.2}{8.4} & \res{97.0}{2.8} \\
            $\rhd$ & \uniqameta & 5 & \res{\bf{90.5}}{5.9} & \res{\bf{93.1}}{4.4} & \res{\bf{81.7}}{5.2} & \res{\bf{76.2}}{7.1} & \res{\bf{98.0}}{2.0} \\ 
        \bottomrule
    \end{tabular}
    \end{subtable}
\end{table}

We compare to these methods using the datasets in the \benchmark benchmark to establish the quality of our model.
Since we constructed our benchmark from disjoint subsets of datasets evaluated in each of these prior works (\S\ref{sec:format}), we compare each method with its corresponding subset of datasets.
Each of these prior works evaluates their methods using different experimental setups (classes, number of episodes, shots) than our benchmark and was not designed to handle \benchmark's challenging episode characteristics like class imbalance.  To enable fair comparison, we test \uniqa on the exact data splits released by the authors when available (H-SMLMT and LM-BFF).
For DS, we sample (balanced) episodes using our framework after matching their test settings (number of shots and classes, class splits, etc.) and reproduce their reported results to within 1\% absolute difference using their model code; we use these episodes for our experiments. The results in Table~\ref{tab:res-compare} show that %
\uniqameta outperforms both H-SMLMT and DS meta-learning approaches by relatively large margins, while achieving competitive results compared with LM-BFF.
Note that \uniqa's strong results are without   meta-learning approaches, extensive prompt-engineering, or per-episode hyperparameter search.

\begin{table}
\centering
    \caption{
    Mean accuracy of \uniqa and \uniqameta on \benchmark benchmark in zero and \fewshot setups.\vspace{-8pt} \\
    }
    \label{tab:res}
\setlength{\tabcolsep}{3.3pt}
\footnotesize
\begin{tabular}{@{}lrrrrrrrrrrc@{}}
\\
\toprule
 &  \multicolumn{3}{c}{Zero-shot} &  &  & \multicolumn{3}{c}{\Fewshot} & & \\
 \cmidrule(l{6pt}r{6pt}){2-4} 
 \cmidrule(l{6pt}r{6pt}){7-9}
& Class & Domain & Task & Overall & & Class & Domain & Task & Overall & & $\Delta_{\text{few}}$ (Overall) \\
\midrule
\uniqa     & 59.5 & 67.9 & 36.6 & 56.5 & & 75.8 & 72.4 & 54.3 & 69.3 & & $+$12.8 \\
\uniqameta & 75.6 & 87.6 & 41.1 & 71.0 & & 80.2 & 86.8 & 62.4 & 77.9 & & $+$6.9 \\
$\Delta_{\text{meta}}$ & $+$16.2 & $+$19.7 & $+$4.5 & $+$14.5 & & $+$4.3 & $+$14.4 & $+$8.1 & $+$8.6 & \\
\bottomrule
\end{tabular}

\end{table}

\paragraph{Evaluating \uniqa on the \benchmark benchmark}

Having established \uniqa as a strong model comparable to recent, state-of-the art techniques, we present its results on the final version of our benchmark (with class imbalance, etc.).
From Table~\ref{tab:res}, we observe three findings. First, pretraining is an effective technique for infusing an NLP model with the ability to perform \fewshot generalization even without any meta-training, as \uniqa is able to score $\Delta_{\text{few}}=+12.8$ higher when provided few rather than zero examples. Second, by comparing \uniqameta and \uniqa, we see that meta-training has a substantial impact on zero-shot performance ($\Delta_{\text{meta}}=+14.5$), but its benefit, while still substantial, is less in the \fewshot setting ($\Delta_{\text{meta}}=+8.6$). Third, while meta-training adds roughly the same benefit to zero and \fewshot performance for both domain and task transfer settings, meta-training disproportionately benefits zero-shot class transfer ($\Delta_{\text{meta}} = +16.2$) over \fewshot class transfer ($\Delta_{\text{meta}} = +4.3$). Such observations are made possible through unified evaluation and comparison across different transfer types. The full \benchmark benchmark results broken down by individual datasets are in Appendix~\ref{sec:appendix:results}.

\section{Limitations and Future Work} 
\label{sec:limitations}
While the initial \benchmark benchmark is focused on classification tasks, we aim to use our benchmark creation toolkit (\S\ref{sec:framework}) to incorporate additional task formats like span selection or text generation.
Furthermore, the benchmark currently only supports English language tasks; to study {\em language transfer}, we aim to incorporate new datasets using our toolkit.
Adding diverse datasets has its own challenges; while we've selected datasets for our benchmark based on prior work adoption and have attempted to verify their licensing for research use, we were unable to find license details for some datasets (Appendix~\ref{sec:appendix:dataset}).
We believe it is crucial to continually evolve the suite of datasets 
to remain challenging for the best models~\cite{Koh2021WILDSAB} and to tackle real-world challenges~\cite{Alex2021RAFTAR}.

In addition, Sample Size Design (\S\ref{sec:stats}) simulations
currently rely on our own available training estimates. We plan to gather a more representative sample from community leaderboard submissions.

Our public leaderboard could benefit from extended support for detailed comparisons between submissions based on properties of techniques.
For example, approaches may vary in terms of model characteristics (e.g., number of parameters), data and supervision used during pretraining,
amount of compute,
etc.
We encourage reporting all these factors to enable the community to analyze and make progress on important sub-spaces in the overall \fewshot technique design space.

Finally, we believe the benefits of improving few-shot NLP techniques outweigh potential risks, but we acknowledge potential harms associated with language models~\cite{Schwartz2020GreenA,Bender2021OnTD,Solaiman2019ReleaseSA,Carlini2020ExtractingTD}.
Few-shot models learn a task from a few examples but rely heavily on knowledge encoded in the pretrained model. %
Thus, few-shot models are more likely to inherit the biases of the pretrained 
models, compared to more fully supervised models; 
as the community focuses more on few-shot learning, 
it is more important than ever for future pretrained models to be 
careful about biases in the underlying pretraining corpora.

\section{Conclusion}
\label{sec:conclusion}

In this work, we 
unify and bring rigor to \fewshot NLP evaluation.
We formulate the \benchmark Principles, a set of requirements and best practices that enables unified, rigorous, valid, and cost-sensitive measurement.
We advance the principles with new Sample Size Design methodology for optimizing statistical accuracy and precision while keeping costs low.
The \benchmark benchmark is our instantiation of the \benchmark Principles; it employs Sample Size Design and includes four \fewshot transfer settings, zero-shot evaluation, and a public leaderboard with diverse NLP tasks.
We present \uniqa, a prompt-based
model that aligns pretraining and downstream task formats, achieving results competitive with recent \fewshot methods despite using trivial prompt engineering.
Finally, we release an extensible, open-source toolkit (used to train \uniqa and generate the \benchmark benchmark) to support future benchmark creation and \fewshot NLP model training.

\begin{ack}
We would like to thank Chandra Bhagavatula, Matt Gardner, Matt Peters, Doug Downey, Dan Weld, and the four anonymous reviewers for helpful comments, suggestions and feedback.
We would also like to acknowledge the large community effort involved in the creation of the datasets and open-source tools we utilize. %
\end{ack}

\bibliography{biblio}
\bibliographystyle{acl_natbib}

\appendix
\section{Datasets}
\label{sec:appendix:dataset}
\paragraph{Dataset stats, tasks, and transfer types}
Table~\ref{tab:dataset} summarizes the tasks and datasets used for meta-training and meta-testing. 
To enable automated benchmark construction and maximize access, we restrict datasets to those that are freely available for automated download.\footnote{We exclude RCV1 (used by \cite{bao2020}) and MPQA (used by \cite{gao2020}), since they require agreeing to license terms through web forms at download time.}
We include the GLUE tasks used by \citet{bansal-coling20,bansal-emnlp20-self-supervised} for meta-training\footnote{We follow \citet{bansal-coling20} and use the matched+mismatched version of MNLI and exclude WNLI and STS-B from meta-training due to the small training size and regression task format, respectively} and thus exclude GLUE tasks used by \citet{gao2020} from meta-testing. Although \citet{bansal-coling20,bansal-emnlp20-self-supervised} additionally use SNLI for meta-training, we reserve it for meta-testing for comparison to \citet{gao2020} and because NLI is already represented in the meta-training datasets.

\paragraph{Textual labels and licenses for datasets}
     We made CoNLL labels more descriptive from their original PER,ORG,LOC,MISC. For TREC, we used the more readable labels from the manual template in \cite{gao2020}. For readability, Amazon labels are shown without underscores and Amazon and HuffPost capitalization has been removed.
     License information is shown in parentheses.
\begin{itemize}
    \item MR~\citep{pang2005} (license unavailable\footnote{\url{https://www.cs.cornell.edu/people/pabo/movie-review-data/rt-polaritydata.README.1.0.txt}}): {\bf Test:} negative, positive
    \item CR~\citep{hu2004} (license unavailable\footnote{\url{https://www.cs.uic.edu/~liub/FBS/CustomerReviewData.zip}}): {\bf Test:} negative, positive
    \item Subj~\citep{pang2004} (license unavailable\footnote{\url{https://www.cs.cornell.edu/people/pabo/movie-review-data/subjdata.README.1.0.txt}}): {\bf Test:} objective, subjective
    \item TREC~\citep{voorhees2000} (license unavailable\footnote{\url{https://cogcomp.seas.upenn.edu/Data/QA/QC/}}): {\bf Test:} description, entity, expression, human, location, number
    \item FewRel~\citep{han2018} (MIT License\footnote{\url{https://huggingface.co/datasets/few_rel}}): {\bf Train:} applies to jurisdiction, architect, child, competition class, constellation, contains administrative territorial entity, country, country of citizenship, country of origin, crosses, father, field of work, followed by, follows, genre, has part, head of government, headquarters location, heritage designation, instance of, instrument, league, licensed to broadcast to, located in or next to body of water, located in the administrative territorial entity, located on terrain feature, location, location of formation, manufacturer, member of, member of political party, military branch, military rank, mother, mountain range, mouth of the watercourse, movement, notable work, occupant, occupation, operating system, operator, owned by, part of, participant, participant of, participating team, place served by transport hub, position held, position played on team / speciality, record label, religion, residence, said to be the same as, sibling, sport, sports season of league or competition, spouse, subsidiary, successful candidate, taxon rank, tributary, voice type, winner, work location;
    {\bf Val:} developer, director, original network, performer, publisher;
    {\bf Test:} after a work by, characters, composer, distributor, language of work or name, main subject, nominated for, original language of film or TV show, platform, screenwriter
    \item HuffPost~\citep{misra2018} (CC0: Public Domain\footnote{\url{https://www.kaggle.com/rmisra/news-category-dataset}}): {\bf Train:} arts, arts \& culture, black voices, comedy, culture \& arts, fifty, food \& drink, good news, green, impact, latino voices, media, money, parenting, religion, sports, style, the worldpost, travel, women;
    {\bf Val:} crime, queer voices, science, weird news, worldpost;
    {\bf Test:} business, college, divorce, education, entertainment, environment, healthy living, home \& living, parents, politics, style \& beauty, taste, tech, weddings, wellness, world new
    \item CoNLL~\citep{sang2003} (license for research by Reuters\footnote{\url{https://www.clips.uantwerpen.be/conll2003/ner/}}): {\bf Test:} location, organization, other, person
    \item SNLI~\citep{bowman2015} (Creative Commons Attribution-ShareAlike 4.0 International License\footnote{\url{https://huggingface.co/datasets/snli}}): {\bf Test:} contradiction, entailment, neutral
    \item SciTail~\citep{khot2018} (license unavailable\footnote{\url{https://allenai.org/data/scitail}}): {\bf Test:} entailment, neutral
    \item Amazon~\citep{he2016} (license unavailable\footnote{\url{http://jmcauley.ucsd.edu/data/amazon/}}): {\bf Train:} automotive, baby, beauty, cell phones and accessories, grocery and gourmet food, health and personal care, home and kitchen, patio lawn and garden, pet supplies, sports and outdoor;
    {\bf Val:} apps for android, cds and vinyl, digital music, toys and games, video games;
    {\bf Test:}  amazon instant video, books, clothing shoes and jewelry, electronics, kindle store, movies and tv, musical instruments, office products, tools and home improvement
    \item 20News~\citep{lang1995} (license unavailable\footnote{\url{https://huggingface.co/datasets/newsgroup}}): {\bf Train:} rec.autos, rec.motorcycles, rec.sport.baseball, rec.sport.hockey, sci.crypt, sci.electronics, sci.med, sci.space;
    {\bf Val:} comp.graphics, comp.os.ms-windows.misc, comp.sys.ibm.pc.hardware, comp.sys.mac.hardware, comp.windows.x;
    {\bf Test:} alt.atheism, misc.forsale, soc.religion.christian, talk.politics.guns, talk.politics.mideast, talk.politics.misc, talk.religion.misc
    \item Reuters~\citep{lewis1997} (license for research by Reuters\footnote{\url{https://kdd.ics.uci.edu/databases/reuters21578/README.txt}}): {\bf Train:} acq, alum, bop, cocoa, coffee, copper, cotton, cpi, crude, earn, gnp, gold, grain, interest, ipi;
    {\bf Val:} iron-steel, jobs, livestock, money-fx, money-supply;
    {\bf Test:} nat-gas, orange, reserves, retail, rubber, ship, sugar, tin, trade, veg-oil, wpi
    \item CoLa~\citep{warstadt2019} (released under fair use\footnote{\url{https://nyu-mll.github.io/CoLA/}}): {\bf Val:} acceptable, unacceptable
    \item MNLI~\citep{williams2018} (multiple licenses\footnote{\url{https://www.aclweb.org/anthology/N18-1101.pdf}}): {\bf Train/Val:} contradiction, entailment, neutral
    \item MRPC~\citep{dolan2005} (license unavailable\footnote{\url{https://www.microsoft.com/en-us/download/details.aspx?id=52398}}): {\bf Train/Val:} equivalent, not\_equivalent
    \item QNLI~\citep{rajpurkar2016} (CC BY-SA 4.0\footnote{\url{https://rajpurkar.github.io/SQuAD-explorer/}}): {\bf Train/Val:} entailment, not\_entailment
    \item QQP~\citep{wang2018} (non-commercial use\footnote{\url{https://www.kaggle.com/quora/question-pairs-dataset}}): {\bf Train/Val:} duplicate, not\_duplicate 
    \item RTE~\citep{dagan2005,bar-haim2006,giampiccolo2007,bentivogli2009} (license unavailable\footnote{\url{https://gluebenchmark.com/}}): {\bf Train/Val:} entailment, not\_entailment
    \item SST-2~\citep{socher2013} (license unavailable\footnote{\url{https://nlp.stanford.edu/sentiment/}}): {\bf Train/Val:} negative, positive
    \item WNLI~\citep{levesque2011} (CC BY 4.0\footnote{\url{https://cs.nyu.edu/~davise/papers/WinogradSchemas/WS.html}}): {\bf Val:} entailment, not\_entailment
\end{itemize}

\begingroup
\setlength{\tabcolsep}{3pt} %
\begin{table*}
    \centering
    \footnotesize
    \caption{\benchmark datasets. Use in prior few-shot evaluation marked indicated with *~\citep{gao2020}, $\dagger$~\citep{bansal-coling20}, and $\ddagger$~\citep{bao2020}. $|\mathcal{Y}_\mathrm{val}|=(k)$ parentheses indicate that the same classes are reused between training and validation. The notation \{$i$:$j$\} is used to denote the set of all integers between $i$ and $j$, inclusive. ``class." and ``doc." are shorthand for ``classification" and document". The ``--''
    indicates that the corresponding dataset is not used for a certain phase, for example, CoLa and WNLI are  only used for meta-validation.  
    }
    \label{tab:dataset}    
    \begin{tabular}{@{}lrrrrrrr@{}}
    \toprule
    {\bf Task Type} & {\bf Dataset} & $|\mathcal{Y}_{\mathrm{train}}|$ & $|\mathcal{Y}_{\mathrm{val}}|$ & $|\mathcal{Y}_{\mathrm{test}}|$ & $|\Ytest|$/ep. & {\bf \#test ex.} & {\bf Transfer} \\ 
    \hline
    \multicolumn{8}{c}{Single-sentence tasks} \\
    \hline
    sentiment & MR*      & -- & -- & 2 & \{2\} & 10662 & Domain \& Pretrain \\
    sentiment & CR*      & -- & -- & 2 & \{2\} & 1708 & Domain \& Pretrain \\
    subjectivity & Subj*    & -- & -- & 2 & \{2\} & 10000 & Task \& Pretrain \\
    question class. & TREC*    & -- & -- & 6 & \{6\} & 500 & Task \& Pretrain  \\
    entity typing & CoNLL$\dagger$  & -- & -- & 4 & \{4\} & 5648 & Task \& Pretrain \\
    relation class. & FewRel$\ddagger$  & 65 & 5 & 10 & \{5:10\} & 7000 & Class \& Pretrain  \\
    news headline topic & HuffPost$\ddagger$& 20 & 5 & 16 & \{5:10\} & 113957 & Class \& Pretrain  \\
    sentiment & SST-2 $\dagger$ & 2 & (2) & -- & -- & -- & --\\
    acceptability & CoLa$\dagger$ & -- & 2 & -- & -- & --  & Task \& Pretrain\\
    \hline
    \multicolumn{8}{c}{Sentence-pair tasks} \\
    \hline
    NLI & SNLI*    & -- & -- & 3 & \{3\} & 9842 & Domain \& Pretrain  \\
    NLI & SciTail$\dagger$ & -- & -- & 2 & \{2\} & 2126 & Domain \& Pretrain \\
    NLI & MNLI$\dagger$ & 3 & (3) & -- & -- & --  & --\\
    QA/NLI & QNLI$\dagger$ & 2 & (2) & -- & -- & -- & --\\
    NLI & RTE$\dagger$ & 2 & (2) & -- & -- & -- & --\\
    paraphrase & MRPC$\dagger$ & 2 & (2) & -- & -- & -- & --\\
    paraphrase & QQP$\dagger$ & 2 & (2) & -- & -- & --  & --\\
    NLI & WNLI & -- & 2 & -- & -- & -- & Domain \& Pretrain \\
        \hline
    \multicolumn{8}{c}{Document tasks} \\
    \hline
    review product  & Amazon$\ddagger$ & 10 & 5 & 9 & \{5:9\} & 9000 & Class \& Pretrain  \\
    informal doc. topic  & 20News$\ddagger$ & 8 & 5 & 7 & \{5:7\} & 6021 & Class \& Pretrain  \\
    document topic  & Reuters$\ddagger$ & 15 & 5 & 11 & \{5:10\} & 835 & Class \& Pretrain  \\
    \bottomrule
    \end{tabular}
\end{table*}
\endgroup

\section{Sample Size Simulations}
\label{sec:appendix:simulation}

We describe how we performed the simulations described in \S\ref{sec:stats}.

\paragraph{Relating $C$, $|\Etest|$ and $\overline{|\Dtest|}$} The cost of meta-testing on \benchmark for a given dataset is the sum of the cost of both \fewshot and zero-shot evaluations:

\begin{align*}
    C &=  C_{\text{few}} + C_{\text{zero}} \\
      &= \left( C_{\text{few}}^E |\Etest| + C_{\text{few}}^I |\Etest| \overline{|\Dtest|} \right) + \left( C_{\text{zero}}^E |\Etest| + C_{\text{zero}}^I |\Etest| \overline{|\Dtest|} \right) \\
      &= |\Etest| \left( (C_{\text{few}}^E + C_{\text{zero}}^E) + \overline{|\Dtest|} (C_{\text{few}}^I + C_{\text{zero}}^I) \right)
\end{align*}

where $C_{\text{few|zero}}^E$ is (average) time spent per-episode during model setup and training, $C_{\text{few|zero}}^I$ is (average) time spent per-episode per-test-instance on evaluation.  We estimate these quantities on a single Titan RTX-8000 GPU with 48Gb memory by conducting meta-testing runs with the \uniqa model (300 steps in \fewshot setting) across all datasets in \benchmark with arbitrary choices for $|\Dtest^E|$.  These tended to be around 95--98 sec for $C_{\text{few}}^E$, 1--3 sec for $C_{\text{zero}}^E$, and 0.7--0.11 sec for $C_{\text{few|zero}}^I$. From this, we derived possible $(C,|\Etest|, \overline{|\Dtest|})$ configurations by solving for $\overline{|\Dtest|}$ over grids of $C = 24, 36, \dots, 84$ and $|\Etest| = 5, 15, 30, 45, \dots, 150$.

\paragraph{Simulating confidence intervals} We describe a single simulation run by a given $(C,|\Etest|, \overline{|\Dtest|})$.  First, we need to generate $\overline{|\Dtest|}$ model predictions for every episode $E \in \Etest$. To do this, we assume each episode has a latent episode-specific model accuracy $\mu_{acc}^{(1)},\dots,\mu_{acc}^{(|\Etest|)}$, where each $\mu_{acc}^{(\cdot)}$ is drawn from a Normal distribution with mean $\mu_{acc}$ and variance $\sigma^2_{acc}$.  Here, $\mu_{acc}$ represents the unknown overall model accuracy that is our target of estimation, and $\sigma^2_{acc}$ represents inherent variability in task difficulty across episodes (e.g., due to different number of classes or imbalance).  In our simulations, we set $\sigma_{acc} = 0.05$.
For each episode $E$, we generate prediction outcomes (i.e. correct or incorrect) from a Bernoulli with success probability $\mu_{acc}^E$.  This allows us to compute episode-specific accuracy estimates $\hat{\mu}_{acc}^{(1)}, \dots, \hat{\mu}_{acc}^{(|\Etest|)}$ and finally compute the mean, standard deviation, and (bootstrap) CI across these episodes.  In doing so, a single simulation run represents a possible submission outcome to \benchmark for a given model, and we can obtain the resulting CI's width and verify whether it contains the true model accuracy $\mu_{acc}$.

\section{Prompts}
\label{sec:prompts}
We use the following prompts for \benchmark\ benchmark tasks based on the input type: 
\begin{table}[H]
\footnotesize
\begin{tabular}{p{\textwidth}}
\begin{itemize}[leftmargin=6pt]
    \item Single text classification: 
    
    \texttt{Topic?\textbackslash\textbackslash n  (A) Class1 (B) Class2 (C) Class3 \textbackslash\textbackslash n \textit{The document}}
    \item Sentence-pair classification:
    
    \texttt{\textit{Sentence 1} Is \textit{Sentence 2}?\textbackslash\textbackslash n (A) Yes (B) No (C) Maybe }
    \item Relation classification:
    
    \texttt{
    \textit{mention-1} to \textit{mention-2}? \textbackslash\textbackslash n (A) Class1 (B) Class2 (C) Class3  \textbackslash\textbackslash n
    \textit{Some text \#mention-1\# some text *mention-2* some text}.}
    \item Entity recognition:
    
    \texttt{
    \textit{What is the type of the entity between the \# marks}? \textbackslash\textbackslash n (A) Class1 (B) Class2 (C) Class3  \textbackslash\textbackslash n
    \textit{Some text \#mention-1\# some text}.}    
\end{itemize}
\end{tabular}
\end{table}

The format of question, followed by the document followed by answer choices, as well as the use of the special delimiter of \texttt{\textbackslash\textbackslash n} is according to UnifiedQA's original pretraining.
We follow \cite{gao2020}'s format of NLI for sentence pair tasks and T5 \cite{Raffel2020ExploringTL} for relation classification.

\section{Baseline Models}
\label{sec:appendix:baselines}
This section briefly describes the baselines we use for comparison. 

\textit{LM-BFF \cite{gao2020}} is a language model prompt-based fine-tuning method with extensive automated and manual approaches for prompt generation. It also uses a strategy for dynamically and selectively incorporating demonstrations into each context which is an extension to GPT-3's in-context learning technique \cite{brown2020}.

\textit{Distributional Signatures (DS) \cite{bao2020}} A meta-learning method designed for class transfer. DS uses lexical ``distributional signatures,'' characteristics of the underlying word distributions to transfer attention patterns across tasks within a meta-learning
framework.

\textit{SMLMT \cite{bansal-emnlp20-self-supervised}} A self-supervised approach for domain and task transfer. SMLMT creates the target task distribution from a large set of unlabeled sentences used within a meta-learning framework for optimal transfer. We compare with the strongest model variant in this paper, Hybrid-SMLMT which is trained on both self-supervised and supervised tasks.

\begin{table}
\centering
    \caption{
    Full results table with all the stats. CI-low and CI-up are the lower and upper 95\% bootstrap confidence intervals (of the mean), and CI-sem is the symmetric 95\% standard error-based confidence interval.
    }
    \label{tab:res_full_stats}
\setlength{\tabcolsep}{3.3pt}
\footnotesize
\begin{tabular}{@{}lllrrrrrrrrrrrrr@{}}
\toprule
 \multirow{2}{*}{Shot}  &  \multirow{ 2}{*}{Model}    &  &  \multicolumn{5}{c}{Class Transfer} & \multicolumn{4}{c}{Domain Transfer}  & \multicolumn{3}{c}{Task Transfer} \\
\cmidrule(l{6pt}r{6pt}){4-8} \cmidrule(l{6pt}r{6pt}){9-12} \cmidrule(l{6pt}r{6pt}){13-15}
  &   & Stat & Amzn            & FRel            & HufP            & 20N             & Reut            
  & CR              & MR              & SciT            & SNLI      
  & CNLL            & Subj            & TREC            \\ 
\midrule

Zero & \uniqa  &  \begin{tabular}{@{}l@{}} Mean \\ Stdev \\ CI-low \\ CI-up \\ CI-sem \end{tabular} & \begin{tabular}{@{}r@{}}69.9 \\ 7.2 \\ 1.45 \\ 1.50 \\ 1.50\end{tabular} &  \begin{tabular}{@{}r@{}}52.5 \\ 9.7 \\ 2.02 \\ 1.85 \\ 2.02\end{tabular} &  \begin{tabular}{@{}r@{}}46.9 \\ 10.1 \\ 2.05 \\ 2.15 \\ 2.10\end{tabular} &  \begin{tabular}{@{}r@{}}43.7 \\ 10.2 \\ 1.95 \\ 2.13 \\ 2.12\end{tabular} &  \begin{tabular}{@{}r@{}}84.3 \\ 6.0 \\ 1.33 \\ 1.24 \\ 1.26\end{tabular} &  \begin{tabular}{@{}r@{}}85.4 \\ 4.7 \\ 0.95 \\ 0.92 \\ 0.99\end{tabular} &  \begin{tabular}{@{}r@{}}77.1 \\ 3.3 \\ 0.67 \\ 0.64 \\ 0.69\end{tabular} &  \begin{tabular}{@{}r@{}}56.4 \\ 3.2 \\ 0.62 \\ 0.62 \\ 0.66\end{tabular} &  \begin{tabular}{@{}r@{}}52.6 \\ 2.5 \\ 0.53 \\ 0.51 \\ 0.52\end{tabular} &  \begin{tabular}{@{}r@{}}31.6 \\ 3.3 \\ 0.68 \\ 0.68 \\ 0.69\end{tabular} &  \begin{tabular}{@{}r@{}}55.2 \\ 3.3 \\ 0.66 \\ 0.68 \\ 0.68\end{tabular} &  \begin{tabular}{@{}r@{}}23.1 \\ 4.7 \\ 0.96 \\ 0.95 \\ 0.97\end{tabular}\\
\hdashline[0.1pt/1pt]\noalign{\vskip 1.0ex}
 Zero & \uniqameta & \begin{tabular}{@{}l@{}} Mean \\ Stdev \\ CI-low \\ CI-up \\ CI-sem \end{tabular} & \begin{tabular}{@{}r@{}}75.6 \\ 8.4 \\ 1.77 \\ 1.81 \\ 1.74\end{tabular} &  \begin{tabular}{@{}r@{}}79.4 \\ 9.2 \\ 1.89 \\ 1.80 \\ 1.91\end{tabular} &  \begin{tabular}{@{}r@{}}68.5 \\ 6.6 \\ 1.26 \\ 1.36 \\ 1.37\end{tabular} &  \begin{tabular}{@{}r@{}}60.0 \\ 8.3 \\ 1.81 \\ 1.76 \\ 1.72\end{tabular} &  \begin{tabular}{@{}r@{}}94.6 \\ 2.1 \\ 0.46 \\ 0.45 \\ 0.44\end{tabular} &  \begin{tabular}{@{}r@{}}93.7 \\ 1.3 \\ 0.27 \\ 0.26 \\ 0.27\end{tabular} &  \begin{tabular}{@{}r@{}}90.8 \\ 1.9 \\ 0.41 \\ 0.41 \\ 0.40\end{tabular} &  \begin{tabular}{@{}r@{}}82.6 \\ 2.2 \\ 0.44 \\ 0.44 \\ 0.46\end{tabular} &  \begin{tabular}{@{}r@{}}83.3 \\ 2.0 \\ 0.39 \\ 0.40 \\ 0.41\end{tabular} &  \begin{tabular}{@{}r@{}}34.8 \\ 1.8 \\ 0.35 \\ 0.40 \\ 0.38\end{tabular} &  \begin{tabular}{@{}r@{}}52.7 \\ 2.9 \\ 0.60 \\ 0.58 \\ 0.60\end{tabular} &  \begin{tabular}{@{}r@{}}35.9 \\ 2.0 \\ 0.41 \\ 0.43 \\ 0.42\end{tabular} & \\

\hdashline[0.1pt/1pt]\noalign{\vskip 1.0ex}

 Few & \uniqa  & \begin{tabular}{@{}l@{}} Mean \\ Stdev \\ CI-low \\ CI-up \\ CI-sem \end{tabular} & \begin{tabular}{@{}r@{}}79.5 \\ 7.5 \\ 1.52 \\ 1.57 \\ 1.56\end{tabular} &  \begin{tabular}{@{}r@{}}79.2 \\ 7.5 \\ 1.54 \\ 1.49 \\ 1.55\end{tabular} &  \begin{tabular}{@{}r@{}}62.8 \\ 7.7 \\ 1.60 \\ 1.55 \\ 1.59\end{tabular} &  \begin{tabular}{@{}r@{}}63.1 \\ 7.8 \\ 1.59 \\ 1.59 \\ 1.62\end{tabular} &  \begin{tabular}{@{}r@{}}94.5 \\ 3.2 \\ 0.67 \\ 0.63 \\ 0.66\end{tabular} &  \begin{tabular}{@{}r@{}}90.1 \\ 6.6 \\ 1.42 \\ 1.26 \\ 1.37\end{tabular} &  \begin{tabular}{@{}r@{}}78.6 \\ 10.5 \\ 2.16 \\ 1.99 \\ 2.18\end{tabular} &  \begin{tabular}{@{}r@{}}64.9 \\ 8.9 \\ 1.67 \\ 1.81 \\ 1.85\end{tabular} &  \begin{tabular}{@{}r@{}}55.8 \\ 9.5 \\ 2.12 \\ 1.75 \\ 1.97\end{tabular} &  \begin{tabular}{@{}r@{}}44.3 \\ 7.9 \\ 1.76 \\ 1.68 \\ 1.65\end{tabular} &  \begin{tabular}{@{}r@{}}60.5 \\ 9.9 \\ 1.99 \\ 2.21 \\ 2.06\end{tabular} &  \begin{tabular}{@{}r@{}}58.1 \\ 7.7 \\ 1.71 \\ 1.63 \\ 1.61\end{tabular} \\

\hdashline[0.1pt/1pt]\noalign{\vskip 1.0ex}

 Few & \uniqameta & \begin{tabular}{@{}l@{}} Mean \\ Stdev \\ CI-low \\ CI-up \\ CI-sem \end{tabular} & \begin{tabular}{@{}r@{}}82.1 \\ 7.0 \\ 1.44 \\ 1.53 \\ 1.46\end{tabular} &  \begin{tabular}{@{}r@{}}87.2 \\ 5.7 \\ 1.20 \\ 1.07 \\ 1.19\end{tabular} &  \begin{tabular}{@{}r@{}}67.9 \\ 7.5 \\ 1.62 \\ 1.49 \\ 1.56\end{tabular} &  \begin{tabular}{@{}r@{}}67.3 \\ 7.8 \\ 1.58 \\ 1.58 \\ 1.61\end{tabular} &  \begin{tabular}{@{}r@{}}96.3 \\ 2.5 \\ 0.55 \\ 0.52 \\ 0.53\end{tabular} &  \begin{tabular}{@{}r@{}}93.2 \\ 2.5 \\ 0.55 \\ 0.48 \\ 0.52\end{tabular} &  \begin{tabular}{@{}r@{}}89.4 \\ 2.8 \\ 0.59 \\ 0.58 \\ 0.58\end{tabular} &  \begin{tabular}{@{}r@{}}83.6 \\ 4.7 \\ 1.00 \\ 0.92 \\ 0.98\end{tabular} &  \begin{tabular}{@{}r@{}}80.9 \\ 4.5 \\ 0.95 \\ 0.89 \\ 0.93\end{tabular} &  \begin{tabular}{@{}r@{}}58.6 \\ 4.7 \\ 1.04 \\ 0.93 \\ 0.97\end{tabular} &  \begin{tabular}{@{}r@{}}68.7 \\ 10.6 \\ 2.30 \\ 2.24 \\ 2.19\end{tabular} &  \begin{tabular}{@{}r@{}}60.0 \\ 6.6 \\ 1.41 \\ 1.39 \\ 1.38\end{tabular} & \\
\bottomrule\vspace{0.2pt}

\end{tabular}

\end{table}

\section{Full Results Statistics}
\label{sec:appendix:results}
Here, we provide full results broken down by dataset in Table \ref{tab:res_full_stats}. We report bootstrap CIs and standard deviations, as recommended by \citet{dhillon2020}.

\begin{table}
\centering
    \caption{
    Mean accuracy (with 95\% standard error-based CIs) of \uniqa and \uniqameta on \benchmark benchmark in zero and \fewshot settings. 
    }
    \label{tab:res_full}
\setlength{\tabcolsep}{3.3pt}
\footnotesize
\begin{tabular}{@{}llrrrrrrrrrrrrr@{}}
\toprule
 \multirow{2}{*}{Shot}  &  \multirow{ 2}{*}{Model}    & \multicolumn{5}{c}{Class Transfer} & \multicolumn{4}{c}{Domain Transfer}  & \multicolumn{3}{c}{Task Transfer}  & \multirow{2}{*}{Avg}  \\
\cmidrule(l{6pt}r{6pt}){3-7} \cmidrule(l{6pt}r{6pt}){8-11} \cmidrule(l{6pt}r{6pt}){12-14}
  &   & Amzn            & FRel            & HufP            & 20N             & Reut            
  & CR              & MR              & SciT            & SNLI      
  & CNLL            & Subj            & TREC           \\ 
\midrule

 \multirow{ 1}{*}{Zero}         & \uniqa    & \res{69.9}{7.2} & \res{52.5}{9.7} & \res{46.9}{10.1} & \res{43.7}{10.2} & \res{84.3}{6.0} & \res{85.4}{4.7} & \res{77.1}{3.3} & \res{56.4}{3.2} & \res{52.6}{2.5} & \res{31.6}{3.3} & \res{55.2}{3.3} & \res{23.1}{4.7} & $56.5$ \\

  & \uniqameta & \res{75.6}{1.7} & \res{79.4}{1.9} & \res{68.5}{1.4} & \res{60.0}{1.7} & \res{94.6}{0.4} & \res{93.7}{0.3} & \res{90.8}{0.4} & \res{82.6}{0.5} & \res{83.3}{0.4} & \res{34.8}{0.4} & \res{52.7}{0.6} & \res{35.9}{0.4} & $71.0$ \\

\hdashline[0.1pt/1pt]\noalign{\vskip 1.0ex}

 \multirow{ 1}{*}{Few}         & \uniqa  &  \res{79.5}{7.5} & \res{79.2}{7.5} & \res{62.8}{7.7} & \res{63.1}{7.8} & \res{94.5}{3.2} & \res{90.1}{6.6} & \res{78.6}{10.5} & \res{64.9}{8.9} & \res{55.8}{9.5} & \res{44.3}{7.9} & \res{60.5}{9.9} & \res{58.1}{7.7} & $69.3$ \\

  & \uniqameta & \res{82.1}{1.5} & \res{87.2}{1.2} & \res{67.9}{1.6} & \res{67.3}{1.6} & \res{96.3}{0.5} & \res{93.2}{0.5} & \res{89.4}{0.6} & \res{83.6}{1.0} & \res{80.9}{0.9} & \res{58.6}{1.0} & \res{68.7}{2.2} & \res{60.0}{1.4} & $77.9$ \\
\bottomrule\vspace{0.2pt}

\end{tabular}

\end{table}

\section{Software Licenses}
\label{sec:appendix:license}

Our code is licensed under Apache License 2.0. Our framework dependencies are:
\begin{itemize}
    \item HuggingFace Datasets\footnote{\url{https://github.com/huggingface/datasets/blob/master/LICENSE}} (Apache 2.0)
    \item Hydra\footnote{\url{https://github.com/facebookresearch/hydra/blob/master/LICENSE}} (MIT License)
    \item Numpy\footnote{\url{https://github.com/numpy/numpy/blob/main/LICENSE.txt}} (BSD 3-Clause "New" or "Revised")
    \item Scipy\footnote{\url{https://github.com/scipy/scipy/blob/master/LICENSE.txt}} (BSD 3-Clause "New" or "Revised")
    \item Pandas\footnote{\url{https://github.com/pandas-dev/pandas/blob/master/LICENSE}} (BSD 3-Clause "New" or "Revised")
    \item Scikit-learn\footnote{\url{https://github.com/scikit-learn/scikit-learn/blob/main/COPYING}} (BSD 3-Clause "New" or "Revised")
    \item Tqdm\footnote{\url{https://github.com/tqdm/tqdm/blob/master/LICENCE}} (MIT License, MPLv2.0)
    \item Click\footnote{\url{https://github.com/kohler/click/blob/master/LICENSE}} (MIT License)
\end{itemize}

Additional dependencies used in \uniqa are:
\begin{itemize}
    \item Transformers\footnote{\url{https://github.com/huggingface/transformers/blob/master/LICENSE}} (Apache 2.0)
    \item PyTorch\footnote{\url{https://github.com/pytorch/pytorch/blob/master/LICENSE}} (Misc)
    \item Pytorch Lightning\footnote{\url{https://github.com/PyTorchLightning/pytorch-lightning/blob/master/LICENSE}} (Apache 2.0)
\end{itemize}

See Appendix~\ref{sec:appendix:dataset} for dataset licenses.

\section{Leaderboard}
\label{sec:appendix:leaderboard}

\begin{figure}
    \centering
    \includegraphics[width=\textwidth]{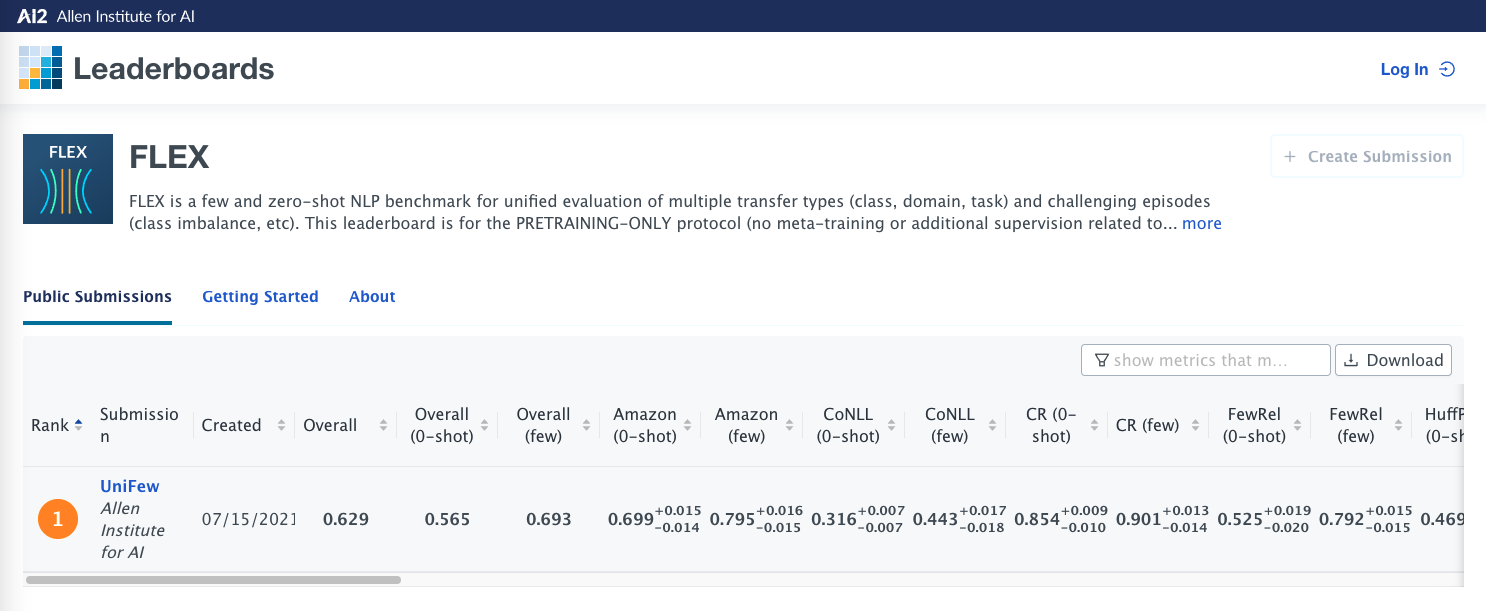}
    \caption{The \benchmark\ public leaderboard for the Pretraining-Only evaluation protocol. A separate leaderboard exists (not shown) for the Meta-Trained protocol.}
    \label{fig:leaderboard}
\end{figure}

Figure~\ref{fig:leaderboard} shows the leaderboard results interface, which displays individual dataset scores with bootstrapped confidence intervals, standard deviations, as well as macro-averaged ``Overall" scores.

\end{document}